\documentclass[letterpaper, 10 pt, conference]{ieeeconf}

\IEEEoverridecommandlockouts

\usepackage{epsfig} 
\usepackage{url}
\usepackage{hyperref}
\usepackage{amsmath} 
\usepackage{amssymb}  
\usepackage{multirow}

\usepackage{multicol}        
\usepackage[bottom]{footmisc}
\usepackage{algorithm}
\usepackage{algorithmic}
\usepackage{epstopdf}

\graphicspath{{figures/}}
\newcommand{\T}{\top} 

\title{\LARGE \bf
Functional Object-Oriented Network for Manipulation Learning
 }
\author{David Paulius, Yongqiang Huang, Roger Milton, William D. Buchanan, Jeanine Sam, and Yu Sun
\thanks{David Paulius, Yongqiang Huang, Roger Milton, William D. Buchanan, Jeanine Sam, and Yu Sun are with the Department of Computer Science and Engineering at the University of South Florida, located in Tampa, Florida, USA. 
Roger Milton, William D. Buchanan, and Jeanine Sam are undergraduate students.
\newline(Contact email: \texttt{davidpaulius,yusun@usf.edu)}}
}

\begin{document}

\maketitle

\thispagestyle{empty}

\pagestyle{empty}

\begin{abstract}
This paper presents a novel structured knowledge representation called the functional object-oriented network (FOON) to model the connectivity of the functional-related objects and their motions in manipulation tasks. 
The graphical model FOON is learned by observing object state change and human manipulations with the objects. U
sing a well-trained FOON, robots can decipher a task goal, seek the correct objects at the desired states on which to operate, and generate a sequence of proper manipulation motions. 
The paper describes FOON's structure and an approach to form a universal FOON with extracted knowledge from online instructional videos.  
A graph retrieval approach is presented to generate manipulation motion sequences from the FOON to achieve a desired goal, demonstrating the flexibility of FOON in creating a novel and adaptive means of solving a problem using knowledge gathered from multiple sources. 
The results are demonstrated in a simulated environment to illustrate the motion sequences generated from the FOON to carry out the desired tasks.

\end{abstract}

\section{Introduction}
\label{sec:intro}
Studies in neuroscience and cognitive science on object affordance \cite{Gibson_1977} indicate that the mirror neurons in the human brain congregate visual and motor responses \cite{Rizz_2004, Rizz_2005, Oztop_2006}.  
Mirror neurons in the F5 sector of the macaque ventral pre-motor cortex fire during both the observation of an interaction with objects and action execution; however, it is important to note that they did not discharge in response to simply observing an object \cite{Di_1992, Gallese_2002}.  
More recently, Yoon et al. \cite{Yoon_2010} studied the affordances associated to pairs of objects positioned for action and found an interesting so-called ``paired object affordance effect.''  
The effect was that the response time by right-handed participants was faster if the two objects were used together when the active object (supposed to be manipulated) was to the right of the other object.
Borghi et al. \cite{Borghi_2012} further studied the functional relationship between paired objects and compared it with the spatial relationship and found that both the position and functional context are important and related to the motion; however, the motor action response was faster and more accurate with the functional context than with the spatial context.
In short, these studies demonstrate that there are strong connections between the observation of objects and the functional motions.
Furthermore, functional relationships between objects are directly associated with the motor actions.  
A comprehensive review of models of affordances and canonical mirror neuron system can be found in \cite{Thill_2013}.

This interesting phenomenon of affordance can be observed in human daily life.  
When humans perform tasks, they pay attention not only to objects and their states but also to object interactions caused by manipulation.
The manipulation reflecting the motor response is tightly associated with both the manipulated object and the interacted object.
Seeking an approach that can connect and model the motion and features of an object in the same framework is considered a new frontier in robotics.
With the boom in learning from demonstration techniques in robotics \cite{Konidaris, Argall, Schaal}, more and more researchers aim to model object features, object affordance, and human action at the same time.
Most of the research builds the relationship between single object features and human action or object affordance \cite{GuptaD, Kjellstrom, Gall,YangAFA15, pieropan2014, Yifan}.
Several studies obtained and used object-action relation without considering many low-level object features; for instance, in \cite{Aksoy} and \cite{Aksoy_2},
concrete object recognition was not considered, and objects were categorized solely according to object interaction sequences.
Objects were segmented in scenes from a number of video sequences, and an undirected semantic graph was used to represent the space interaction relationship between objects.
With a sequence of graphs, their work was able to represent temporal and spatial interactions of objects in an event.
Using these graphs, they constructed an event table and a similarity matrix, and the similarity between two sequences of object interaction events can be obtained according to the matrix.
The objects could further be categorized according to their roles in the interactions, and the obtained semantic graphs might be used to represent robotic tasks.
Jain et al. \cite{jain2009} developed symbolic planning that coupled robot control programs with statistical relational reasoners to arrange objects, such as setting a dinner table, by statistical relational learning.
Yang et al. \cite{yang2014manipulation,Yang_2013_CVPR} proposed a manipulation action tree bank to represent actions of manipulations at multiple levels of abstraction.

Our previous work \cite{SunRAS2013, Ren2013} investigated object categorization and action recognition using an object-object-interaction affordance framework.
We developed an approach to capitalize on the strong relationship between paired objects and interactive motion by building an object relation model and associating it with a human action model in the human-object-object way to characterize inter-object affordance, and thus use the inter-object affordance relationship to improve object and action recognition.
Similar to the mirror neurons in human brains that congregate the visual and motor responses, a novel functional object-oriented network or FOON is presented in this paper that connects interactive objects with their functional motions to represent manipulation tasks.  
The proposed novel FOON focuses on the core of a manipulation task that is determined by both the objects' states and the objects' functional motions, which are represented in a FOON as connected nodes.  
The connections between these nodes represent two-way dependencies, where functional motions depend upon the objects' states and the resulting state depend upon the functional motion.  
A FOON provides structured knowledge not only about the objects and their states but also about the relationship between the functional motions and states.  
From a manipulation goal, a FOON can be searched to find the objects involved, their desired states, and the functional motions to achieve those states.  

\section{Functional Object-Oriented Network}
\label{sec:network}

The proposed FOON is a bipartite network that contains motion nodes and object state nodes. 
In general, an interactive manipulation motion of multiple objects will result in a state change from so-called input objects states to outcome objects states. 
Therefore, we connect the input object state nodes to the outcome object state nodes through the manipulation motion node. 
This arrangement would only allow the object state nodes to be connected to motion nodes and the motion nodes to be connected to object nodes, which thus forms the bipartite network.



\subsection{Nodes of FOON}
The nodes in a bipartite FOON have two types: object-state {\it O} or motion {\it M}.  
In a manipulation task, an object state node {\bf $N_{O}$} represents an object in a certain state, which is either manipulated by a manipulator or is passively interacting with another object. 
Objects may also be seen as containers of other objects, typically ingredients.
These would cover objects such as bowls, pans or ovens which are manipulated with objects within it.
For example, in a cooking task where a person chops a tomato with a knife, both the tomato and the knife can be represented as object state nodes. Initially, the tomato has a state ``whole'' and the knife has a state ``clean''.
After a chopping motion, which is a motion node, the outcome object states are chopped tomato and a dirty knife. 
A motion node {\bf $N_{M}$} contains the type of the manipulation.
From this point, we use \textit{object node} that is short for object state node. 



In a FOON, no two object nodes are the same, or in other words, each object node in the graph is unique in terms of its name and attributes.
However, several motion nodes of the same type can exist at multiple locations in the graph, thus allowing FOON to contain more information than a regular bipartite network.


\subsection{Edges of FOON}
A FOON is identified as a directed graph, as some nodes are the outcomes of the interaction between other nodes.
An edge, denoted as $E$, connects two nodes. 
Edges are drawn from either an object node to a motion node, or vice-versa, but it is important to note that {\it no two objects or two motions are connected to each other}.
The only exception to this is when we transform this bipartite representation into a {\it one-mode projected graph} \cite{Newman}.
These representations are required for network analysis, which will be discussed further in the paper.
In addition, if several object nodes have edges connected to a motion node, it indicates that the objects are interacting with the motion.
If a motion node has edges directed to object nodes, it indicates that the objects are the outcomes of the motion.

A FOON can be called a {\it directed semi-acyclic graph}; this means that there may be some instances of loops where a motion does not necessarily cause a change in an object, as certain objects will remain in the same state.



\begin{figure}[t]
\centering
\includegraphics[width=5cm]{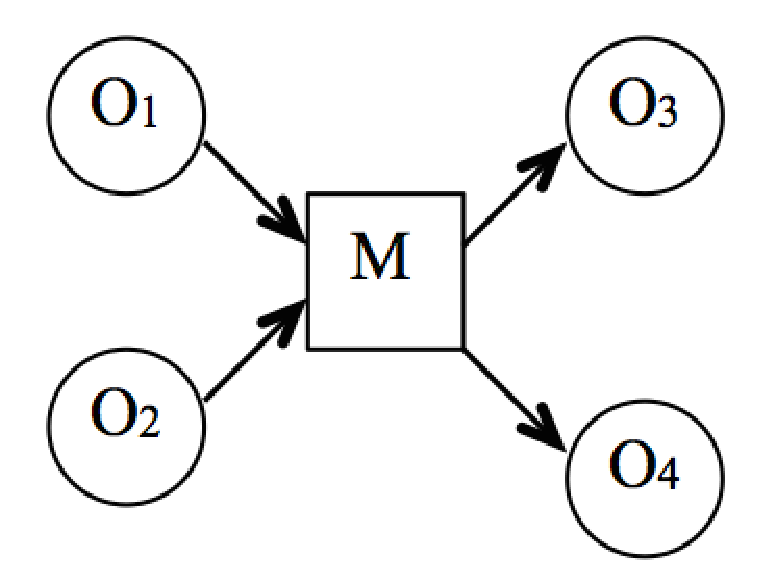}
\caption{A basic functional unit with two input nodes and two output nodes connected by an intermediary single motion node. }
\label{fig:unit}
\end{figure}

\subsection{Functional Unit}

A {\emph {functional unit}} is considered as the minimum learning unit in a FOON.  
It represents the relationship between one or several objects and {\it a single} functional motion associated to the objects. 
In other words, each unit represents a single, atomic action that is part of an activity. 
As shown in Figure \ref{fig:unit}, the object nodes connected with the edges pointing to the functional motion node are called input object nodes, while the object nodes connected with the edges pointing from the functional motion node are called output object nodes.



\subsection{Network Data Structure}
A FOON is represented by conventional graph representations, namely {\it adjacency matrices} and {\it adjacency lists}. 
We use an adjacency matrix to represent the network for its simplicity in representing a digraph and for performing network analysis. 
Each node is represented as a row, and its connectivity to other nodes is represented by columns of the matrix.
An edge from a node $N_{i}$ to $N_{j}$ is denoted by a value of 1, preserving directional details attributed to edges; if two nodes are not connected, then an index has a value of 0.
Accompanying the adjacency matrix is a {\it node list}, which keeps track of all object and motion nodes found in the graph. 
This list is needed to map each node to its row/column representation. 


\section{Learning FOON}
Ideally, a FOON can be automatically trained from observing human activities. 
However, due to the complexity of object, state, and motion recognition, we currently construct many small sets of functional units by labeling instructional videos. 
We manually input these functional units by hand, after which we then merge them together automatically into a single subgraph for each video. 
All subgraphs from each video gathered are then merged into a large network, which we refer to as a \textit{universal FOON}.
Although the functional units are annotated manually, the process of combining the knowledge together as a universal FOON is done algorithmically.
For this reason, the creation of a FOON can thus be seen as a {\it semi-automatic} process.

\subsection{Creating Subgraphs}
\begin{figure}[t]
\centering
\includegraphics[trim={3cm 2.3cm 0cm 3cm},clip,width=\columnwidth]{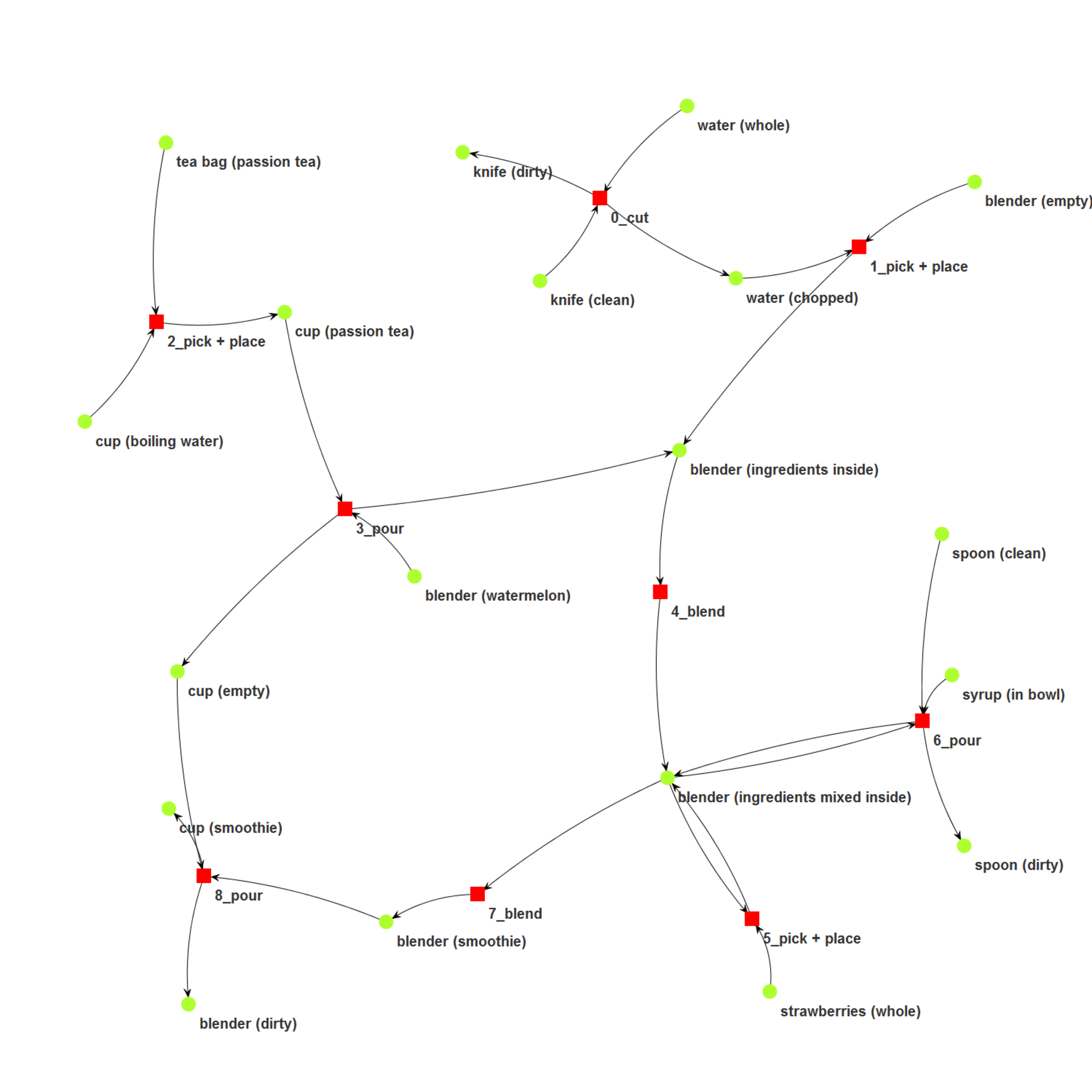}
\caption{A FOON subgraph based on an instructional video on making a watermelon-strawberry smoothie. The green solid circles are object nodes and the red solid squares are motion nodes.
		The object nodes are labeled with object name and their states in parentheses. The motion nodes are labeled with their manipulation motion types. }
\label{fig:garlicbread}
\end{figure}

At the time of this paper, we recruited five volunteers to manually label the input object states, manipulation motion, and output object states in instructional videos through an annotation interface that we have developed.
This interface displays the potential graph that could be made from a given set of functional units. 
The annotations are then converted into functional units with time sequence labels. 
These functional units are then connected and combined into a subgraph automatically using the time sequence labels.
For each video, its FOON subgraph is visualized and verified manually.
Each subgraph provides the essential structured knowledge needed to prepare the dish including objects (ingredients and utensils), their states, and their interactive motions.
Figure \ref{fig:garlicbread} illustrates the FOON subgraph obtained from an online instructional video for preparing a smoothie.

\begin{figure}[t]
\centering
\includegraphics[width= 8.5 cm]{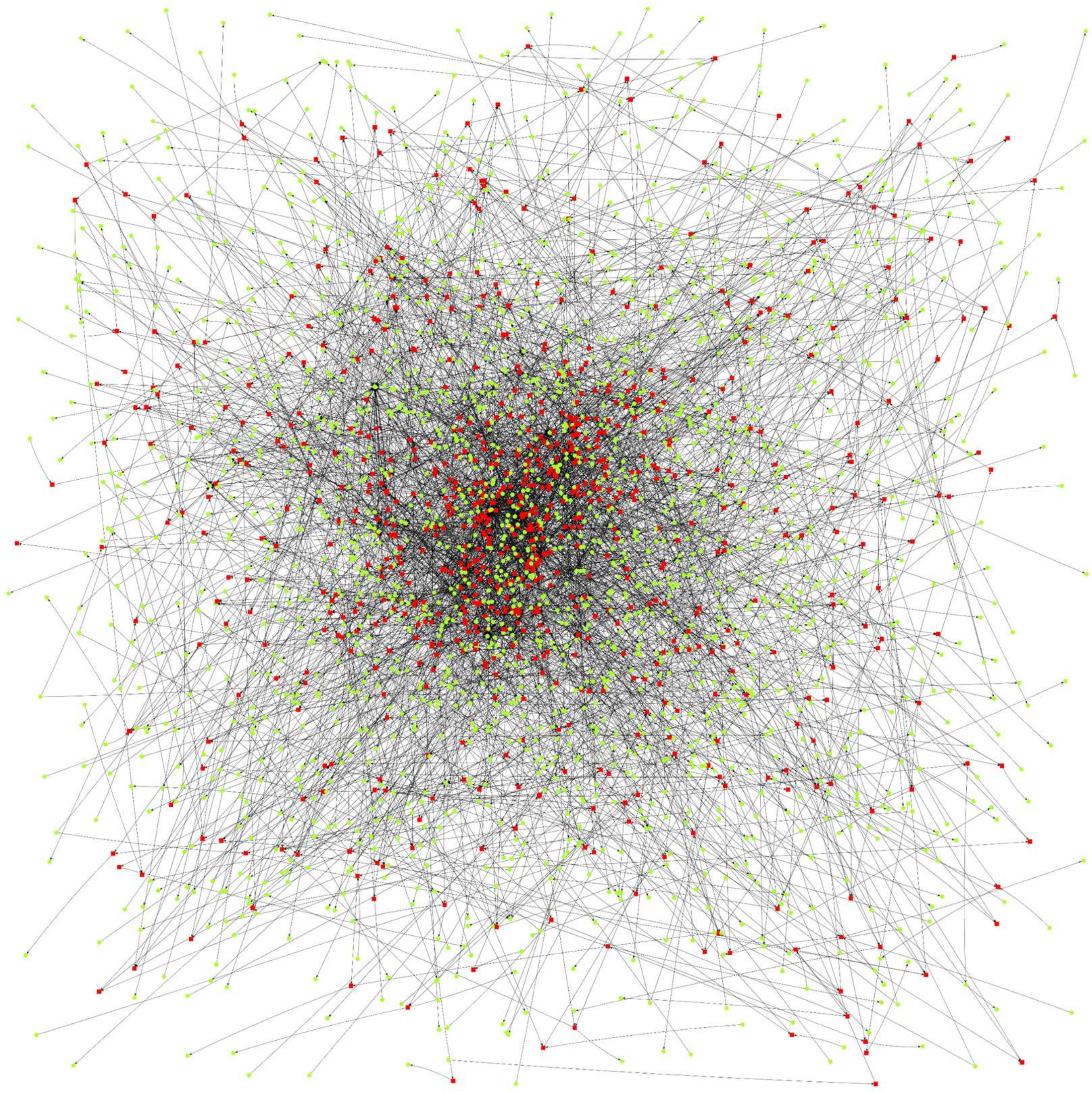}
\caption{Our current universal FOON that is constructed from 60 videos.}
\label{fig:FOON}
\end{figure}

\subsection{Merging Subgraphs}

The universal FOON can be expanded by merging new subgraphs when many videos are processed. The merging algorithm is described as Algorithm 1.

Since subgraphs are individually created by different volunteers, they are prone to inconsistencies from different labels.
Therefore, we developed a parser to pre-process all labeled subgraph files to keep all labeling consistent.
The parser has three main functions: 1) create a main index with a list of all the objects, 2) updating the input file by relabeling all of the objects so they are consistent throughout, and 3) creating a records file that records all changes in any modified files. 
To keep track of all data elements, we create a records file that contains the object name, its old identifier, its new identifier, initial state, final state, file name, and motion.
The parser also finds possible duplicates in objects or motions by using WordNet \cite{WordNet} by comparing the stem word with the current object index.
\begin{algorithm}[t]
\caption{Merge new functional unit in FOON}
\label{merge}
\begin{algorithmic}[H]
\STATE Let $G_{FOON}$ be existing universal FOON
\STATE Let $FU_{new}$ be new functional unit to merge
\\ \vspace{0.3em}\COMMENT{{\it Check if unit is already in FOON:}}
\STATE $found = False$
\FORALL{functional unit $FU_{i}$ in $G_{FOON}$}
\IF {$FU_{i}$ is equal to $FU_{new}$}
\STATE $found = True$
\ENDIF
\ENDFOR
\\ \vspace{0.3em}\COMMENT{{\it If exact match not found, add this unit to FOON:}}\IF {$found$ is not equal to $True$}
\STATE Add $FU_{new}$ to $G_{FOON}$
\STATE Add input nodes $N_{Input}$ to node list
\STATE Add output nodes $N_{Output}$ to node list
\STATE Add motion node $N_{Motion}$ to node list
\ENDIF
\end{algorithmic}
\end{algorithm}

After the nodes are made consistent within all the subgraphs, we run the union operation to merge all subgraphs into a universal FOON graph, one at a time.
At first, the universal FOON is initially empty. 
The union operator first checks if the functional unit is already present in FOON.
Functional units are kept unique, and similar units will have almost the same objects and states found within them.
After this process, objects in the new unit will be added as if they did not exist in the universal node list; if they exist, a reference is made to those existing nodes and then the edges are connected to a new motion node.

So far, we have processed 60 instructional videos on food preparation. Our universal FOON presently has 2169 nodes (broken down into 1229 object nodes made unique by states and 792 instances of 57 possible motion nodes) with 3223 edges after the merging process.
These numbers gradually increase as more subgraphs are continuously being generated.
A low-resolution visualization of the generated universal FOON is shown in Figure \ref{fig:FOON}. 
The high resolution image of the FOON is exceptionally large with a size of more than 200 MB.
The full list of videos and their functional unit subgraphs along with the universal FOON graph are all available for download at \cite{foonet}.

\section{Motion Learning}

For the purpose of motion generation, the motion type in the FOON is represented using motion harmonics \cite{Huang2015}, which are extracted from demonstrated data using functional principal
component analysis (fPCA) \cite{ramsay2005}.
We collected manipulation data for motion learning using an OptiTrack 3D motion capture system in our lab (Figure \ref{fig-mocap}).
Although processing an online 2D video can also produce motion data, the acquired data will be 2-dimensional, which is insufficient to generate executable motions in the 3D world.
Therefore, with our 3D motion capture system, we collected the position and orientation of the objects.

Let $X = \{x_1, x_2, ..., x_N\}$ represent the data that includes $N$ trials, where $x_i(d, t)\in\mathbb{R}$ denotes the value of degree $d$ of trial $i$ at time step $t$, $d=1, 2,\dots,D$, $t=1, 2,\dots, T_i$. 
We assume that six degrees are used: $\{x, y, z, \phi, \theta, \psi\}$, in which $x, y, z$ refer to location coordinates and $\phi, \theta, \psi$
refer to Euler angles.
Unlike location coordinates, Euler angles invariably have ranges, such as $[-\pi, \pi]$. 
For simplicity, we assume the range is $[-1, 1]$.
Thus, to facilitate optimization for motion generation, we apply inverse hyperbolic tangent on the angle trajectory data:

\begin{figure}[t]
	\includegraphics[width=\linewidth]{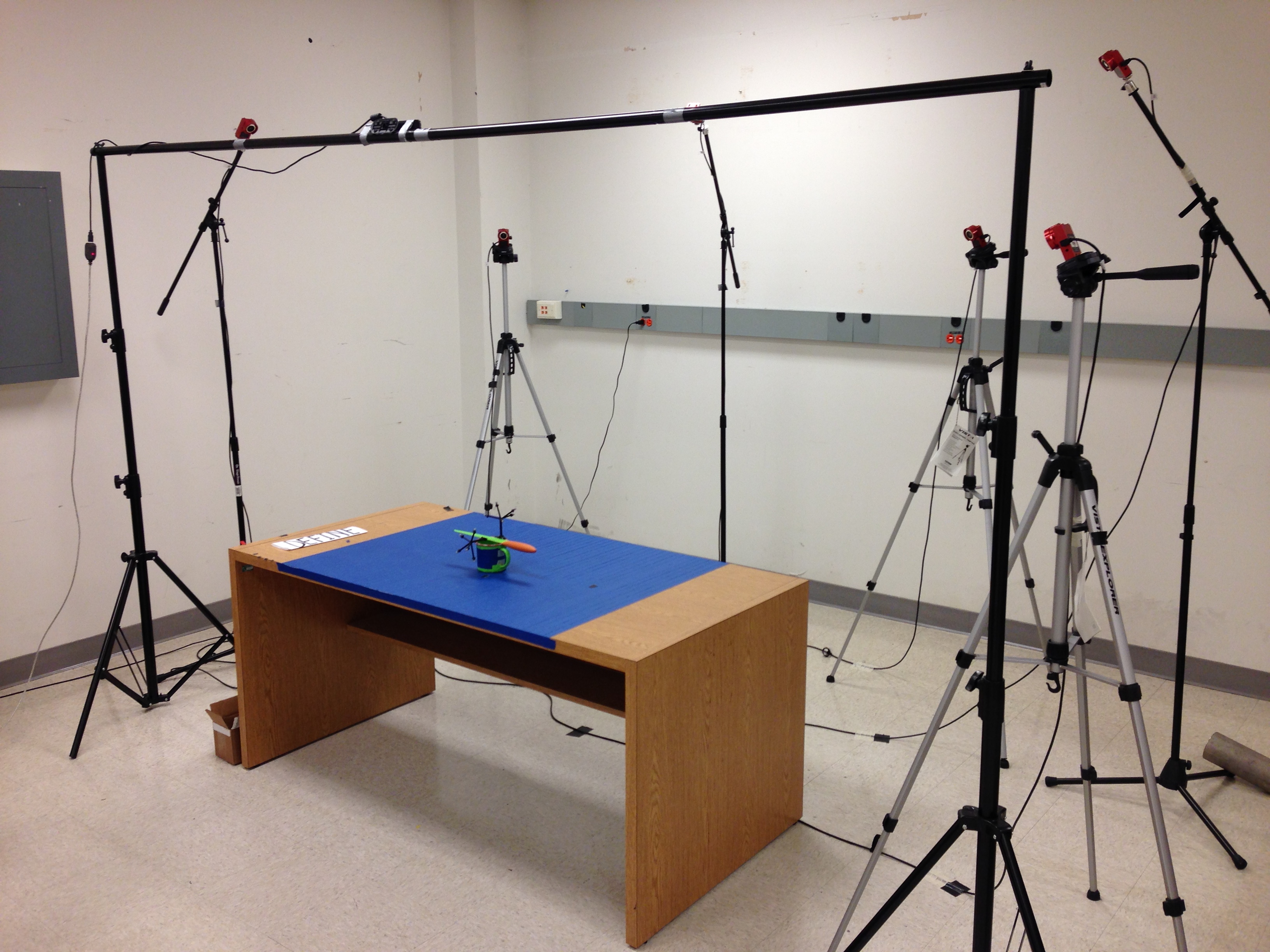}
	\caption{The OptiTrack motion capture system with which we collect data for motion learning. The system consists of six motion capture cameras on tripods.
			Within the blue area on the desk are two objects with reflective markers attached to them. }
	\label{fig-mocap}
\end{figure}

\begin{equation*}
x_i(d)\leftarrow \text{atanh}(x_i(d)), \quad d \in D_q
\end{equation*}
where $D_q$ represents the set of angular degrees.

We used batch Dynamic Time Warping (DTW) \cite{kassidas_etal1998} to align the trials and obtain $Y = \{y_1, y_2, ..., y_N\}$, where $y_i(d, t)\in\mathbb{R}$ corresponds to $x_i$, $t = 1, 2, \dots, T$, and $T$ is the common trajectory length. Applying fPCA to data $Y$, we obtain the mean trajectory $\bar{g}(d, t)\in\mathbb{R} $, the motion harmonics $g(d, t)\in\mathbb{R}^M$ where $M$ is the number of motion harmonics, and the weights $c_i(d)\in\mathbb{R}^M$, $i=1, 2, \dots, N$. We assume the weights are produced from a Gaussian distribution, whose maximum likelihood parameters equal the sample mean $\mu(d)\in\mathbb{R}^M$ and sample covariance $\Sigma(d)\in{R}^{M\times M}$ of the weights. Thus the Gaussian is denoted by $\mathcal{N}(\mu(d), \Sigma(d))$.

To summarize, the motion demonstrated in the data is represented by
\begin{equation*}
\Theta = \left(g(d, t), \bar{g}(d, t), \mathcal{N}(\mu(d), \Sigma(d))\right), \quad d=1, 2, \dots, D
\end{equation*}
We refer readers to \cite{Huang2015} for more detail of motion harmonics as a motion representation.

\section{Analysis of FOON}


We primarily focus on determining the most central or important nodes in our network. 
The importance of a node is reflected by the frequency at which said node interacts with other nodes. 
This measure of importance is referred to as {\it centrality}, and this is a computed value that is assigned to each node.
There are many ways of computing the centrality, and the measures we have applied to FOON were {\it degree} centrality, {\it eigenvector} centrality and {\it Katz} centrality.
The one-mode projected network is used specifically for centrality analysis on objects to observe the relationship between tools and ingredients used in a FOON.
We refer readers to \cite{Newman} for a more in-depth review on graph theory. 
We can apply the obtained information to our specific application, where we can determine from the object nodes which objects need to be frequently used by the robot, and from the motion nodes which manipulation skills are the most important for the robot to learn and execute with particularly high dexterity.


\subsection {Object Centrality}

\begin{figure}[t]
\centering
\includegraphics[trim={0cm 1.2cm 0cm 0cm},clip,width=\columnwidth]{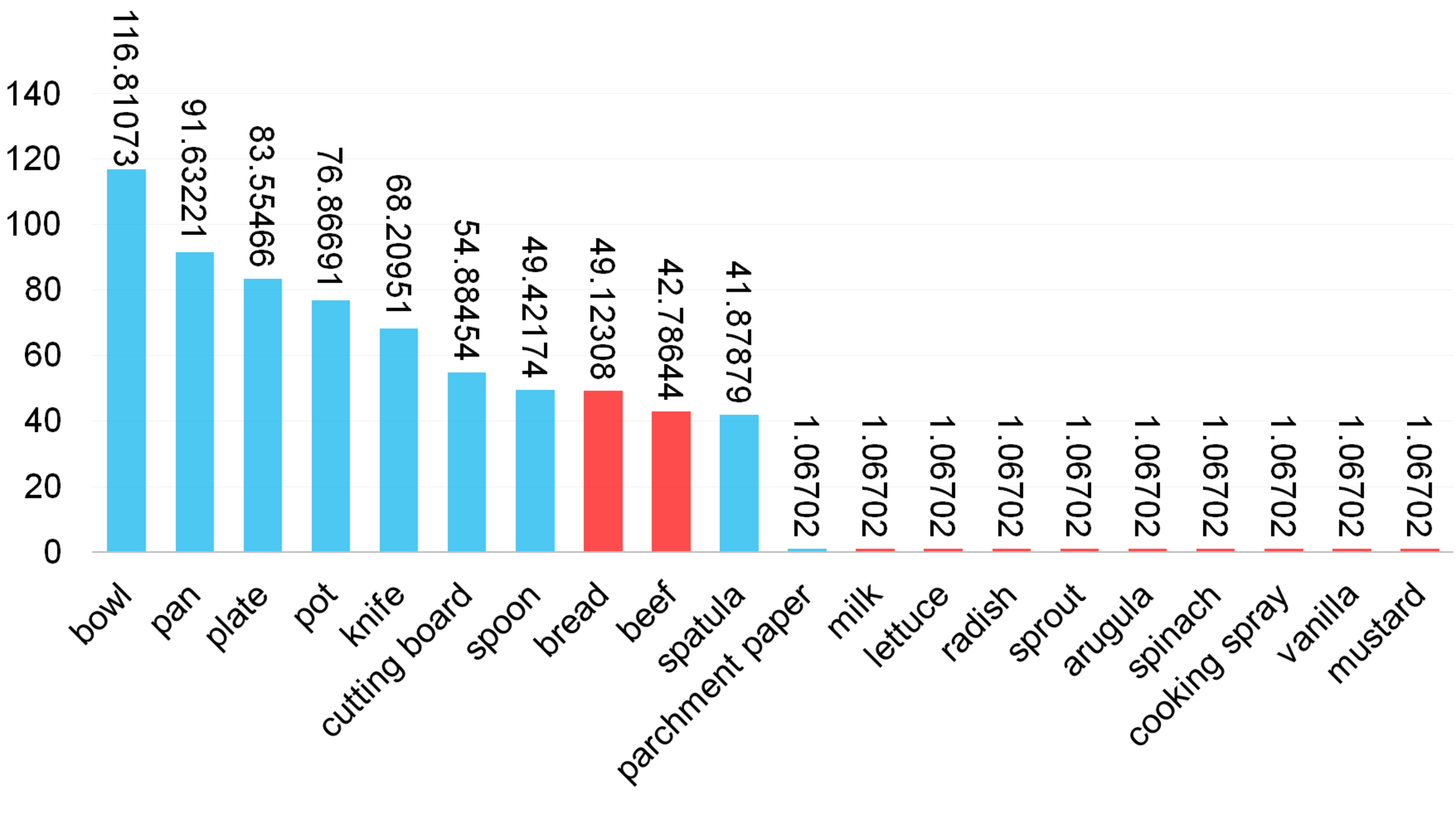}
\caption{Graph showing the objects with the 10 highest and 10 lowest centrality values. The higher the value, the more important a node is. Objects are also classified as utensils (shown in blue) and ingredients (shown in red).}
\label{fig:obgraph}
\end{figure}

We chose to look at the relationship between objects by converting it into a one-mode projected network \cite{Newman}.
The one-mode projection removes all intermediary motion nodes and so all object nodes are directly connected to one another;
in this way, we can also investigate object-object relationships. 
Objects are connected to each other based on the direction of edges  in all functional units in FOON.
Centrality values are not simply integer values, as the computations involve more than counting the node's degree. 
We should not only be concerned with the degree of each node, but we also ought to consider the influence of all nodes connected to every other node.

Our findings for the most important nodes when using these three measures were the same when not considering states; the bowl object
was found to be the most important object with a total of 72 edges. Other objects along with their Katz centrality values are illustrated in Figure \ref{fig:obgraph}.
We can use this information in determining the objects which require the most attention in mastery and skill in manipulating them. The centrality values also let us know which
objects are in high demand in recipes across the entire network; this is important for us to know so that we ensure that these objects are available in our working environment.

\subsection {Motion Frequency}
We also consider the frequency at which objects and motions appear in our network, which we can use for determining the most likely action to occur at a given time and with a
given object. 
We do this by counting the number of instances of each motion belonging to a functional unit that were found in the network.
The most frequent motion observed out of 57 possible motion types is the {\it pick-and-place} motion. This makes sense as there is much translation and movement of objects when preparing meals. 
For us, this means that robots used for cooking tasks should have mastered the pick-and-place motion for different objects.



\begin{figure}[t]
\centering
\includegraphics[width=7cm]{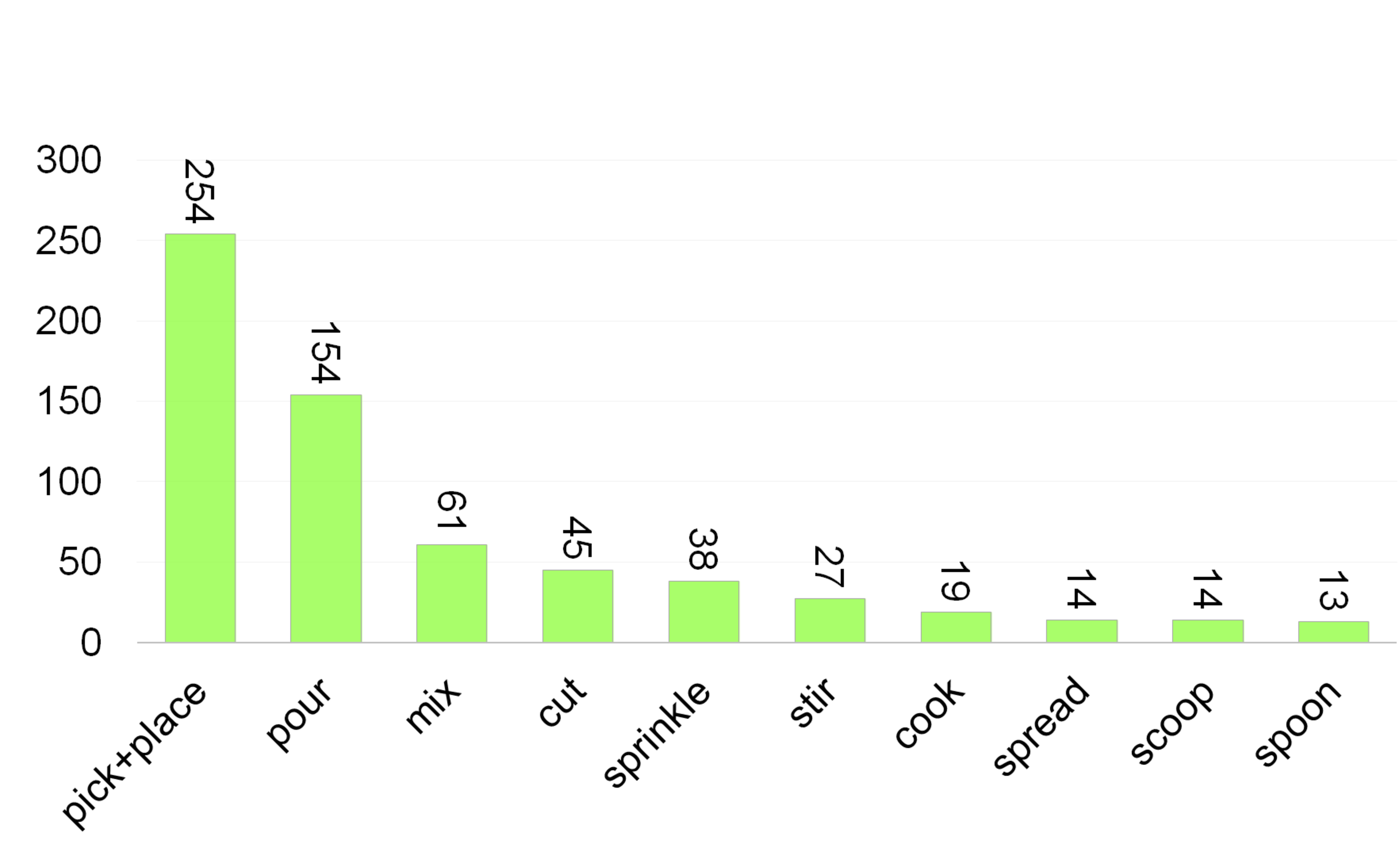}
\caption{Graph showing the top 10 motions observed in our universal FOON (out of 798 motion instances).}
\label{fig:motgraph}
\end{figure}

The next motion found to be frequent is the {\it pour} motion. We believe this is due to the nature of cooking, where items are usually mixed together or put into containers from
other objects. The top 10 motion frequencies can be found in Figure \ref{fig:motgraph}. These values were found after the merging process, and so these are the most frequent nodes after compression of the network.
We can confidently say that these do reflect the reality of cooking in the kitchen.

With these probabilities, we hope to improve our structure to behave more like a typical probabilistic graphical model within the next phases of our project.
The frequencies can be used for compressing FOON even further by possibly removing the need for duplicate motion nodes. 
When paired with the objects, our system would be able to determine the next likely outcome for each object and thus making robot manipulations easier to perform.

%
%
%
%
%
\section{Knowledge Retrieval from a FOON}
The universal FOON can be considered as a structured knowledge representation that can be used for solving manipulation tasks.
Given a desired goal and a set of available objects, formally, there are two steps in generating manipulations from the FOON: 1) retrieving a task tree, and 2) generating the motions needed to accomplish the task.
The retrieval algorithm used is a combination of the breadth-first search and the depth-first search as a specialized application of the branch-and-bound algorithm.

\subsection{Retrieving Task Tree}

Let $N_{Goal}$ be a goal node in the FOON that has been identified as a desired goal or product given to the robot.
The initial task tree $T$ will be empty. 
$T$ will be populated with functional units that make $N_{Goal}$ (i.e. functional units that will output this goal).
To facilitate the search, we require a set of lists: a list of all items available to the robot in the kitchen or its environment ($K$), a list of items we do not know how to make ($S$), and a list of candidate functional units that produces nodes in $S$ ($C$).

Before the search is initiated, we add $N_{Goal}$ to $S$, as it is an object we do not initially know how to make.
We commence the search by removing the first element from $S$, which is denoted as $N_{Head}$, and we search the universal FOON $G_{FOON}$ for all procedures that output $N_{Head}$; we would add these units to list $C$ as potential actions to take. 
Once we have identified all candidate units for $C$, we then search for the ideal functional unit $FU_{i}$ in $C$ that we can execute in its entirety.
This is determined by the availability of objects in each functional unit as listed in $K$. 
In other words, if we find that all input objects $N_{Input}$ in any unit are available in $K$, we can add this unit to $T$ and mark $N_{Head}$ as ``solveable'' or seen.
However, if we cannot find any functional unit that can be executed fully due to missing objects, then we add those items $N_{Input}$ to $S$ as additional sub-goals that need to be met.
We repeat the search for functional units that produce $N_{Head}$ instances as long as there are items in $S$ that we need to know how to make, especially $N_{Goal}$.
We do this process entirely until it has been determined that there is no possible way of solving the manipulation task (due to shortage of items) or until we have found an executable task tree.
We will know that a task tree sequence is found when $N_{Goal}$ has been marked as being ``solveable'', which once again is denoted by its existence in $K$.

\begin{algorithm}[t]
\caption{Task tree retrieval algorithm} \label{retrieval}
\begin{algorithmic}[H]
\STATE Let $N_{Goal}$ be the desired output node
\STATE Let $K$ be list of objects in scene or kitchen
\STATE Let $S$ be list of objects to search
\STATE Let $T$ be final task tree
\STATE Let $N_{Head}$ be head of $S$
\STATE Let $C$ be list of candidate functional units for $N_{Head}$
\STATE Add $N_{Goal}$ to $S$
\vspace{0.3em}
\STATE \COMMENT{{\it Iterate until the goal has been achieved:}}\WHILE{$N_{Goal}$ not in $K$}
\STATE Pop $N_{Head}$ from $S$ 
\STATE \COMMENT{{\it Add units containing head as output:}}
\FORALL{$FU_{i}$ in $G_{FOON}$}
\IF {$N_{Head}$ is $N_{Output}$ of $FU_{i}$}
\STATE Add $FU_{i}$ to $C$
\ENDIF
\ENDFOR
\STATE \COMMENT{{\it Check if inputs to a candidate are available:}}
\FORALL{$FU_{i}$ in $C$}
\FORALL{$N_{Input}$ in $FU_{i}$}
\STATE \COMMENT{{\it Add input as new sub-goal to solve:}}
\IF {$N_{Input}$ not in $K$}
\STATE Add $N_{Input}$ to $S$
\ENDIF
\ENDFOR
\IF {every $N_{Input}$ of $FU_{i}$ is in $K$}
\STATE \COMMENT{{\it Add candidate unit to final task tree:}}
\STATE Add $FU_{i}$ to $T$ 
\STATE Remove all functional units in $C$
\ENDIF
\ENDFOR
\ENDWHILE
\STATE Return $T$
\end{algorithmic}
\end{algorithm}

The searching procedure can also be adjusted to make use of weights that can act as heuristics or constraints on the creation of a task tree.
These heuristics can be a {\it cost} value that is associated with each motion, influencing the selection of functional units which are added to $T$.
We hope to improve the quality of our search through the use of cost values reflecting the complexity of motions or finding the shortest path to accomplishing a goal.

\begin{figure}[t]
\centering
\includegraphics[width=8.5cm]{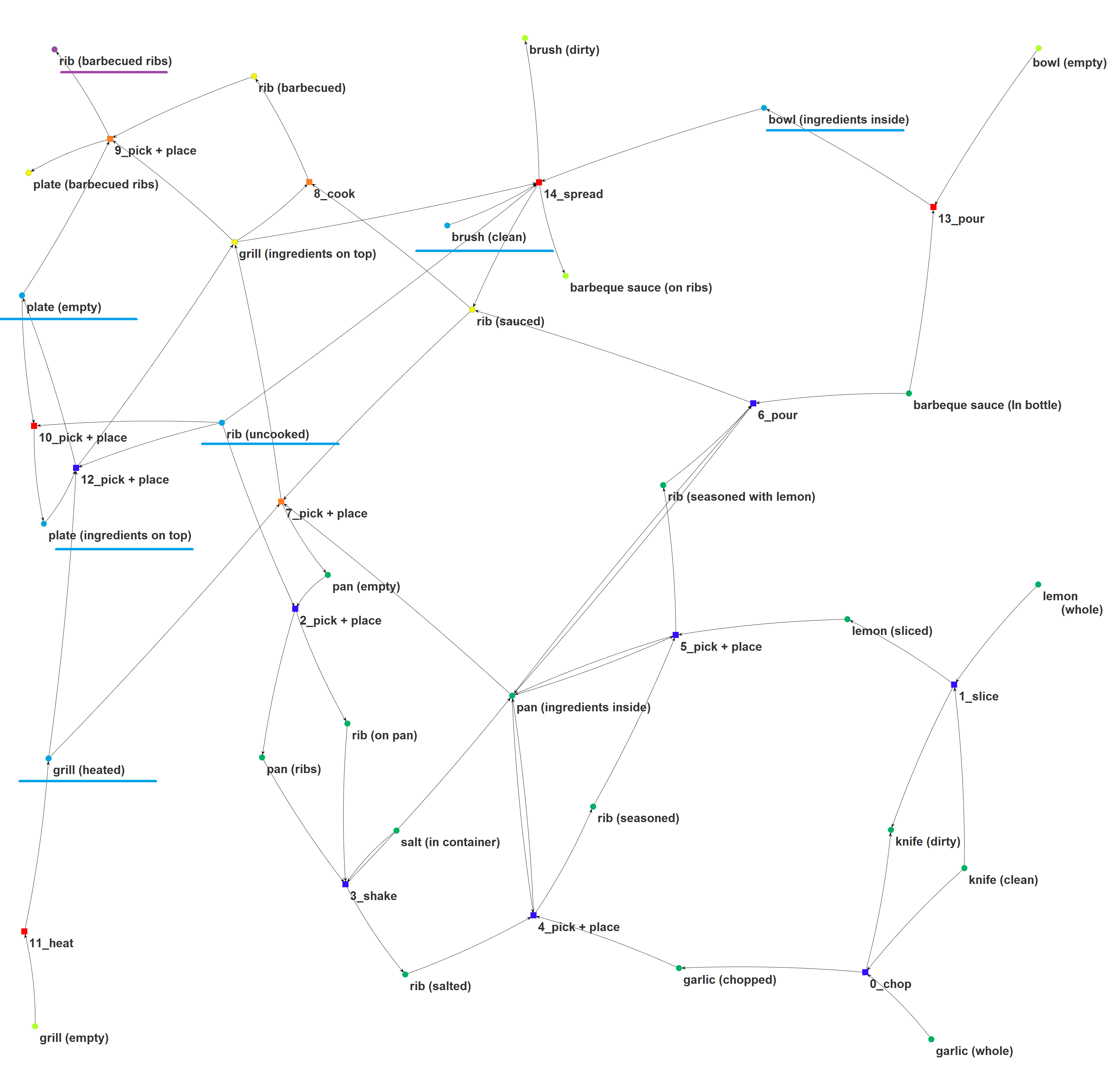}
\caption{Example of a FOON merging two ways of preparing cooked ribs barbecued ribs (node in purple) using available objects (in blue).}
\label{fig:tree}
\end{figure}

\begin{figure}[t]
\centering
\includegraphics[trim={0cm 1cm 0cm 1cm},clip,width=8cm]{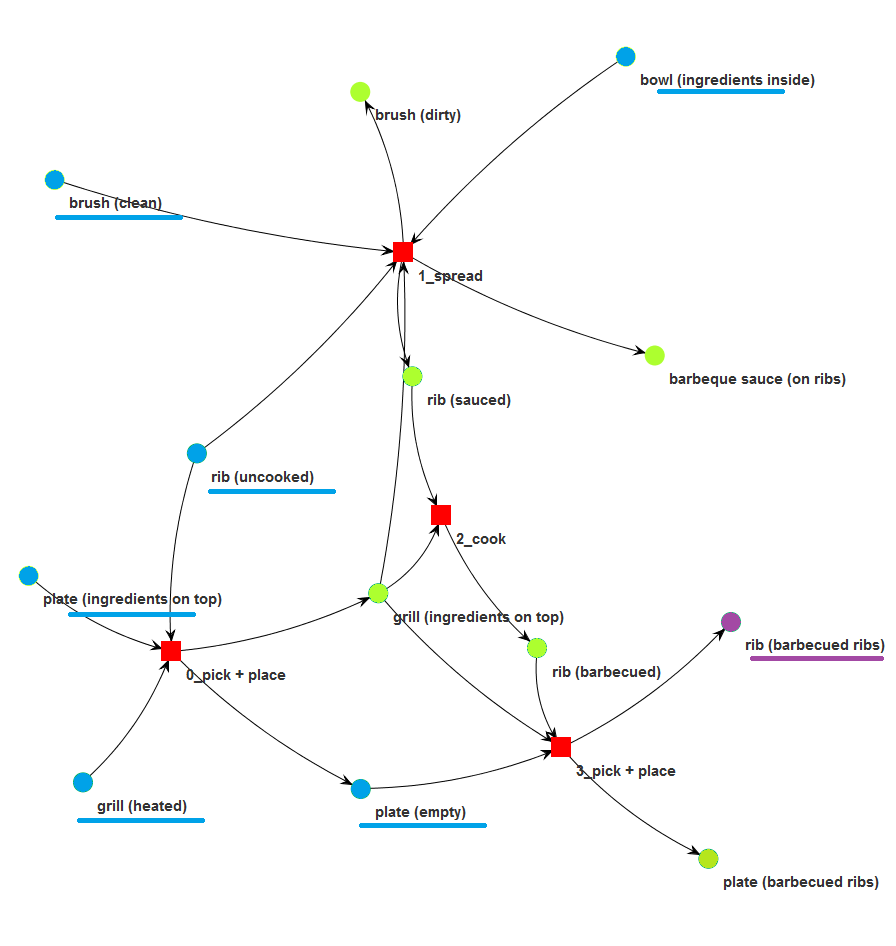}
\caption{Task tree showing the steps needed to prepare barbecued ribs (highlighted in purple) using available objects (in blue) to accomplish goal.
		}
\label{fig:stree}
\end{figure}

We now demonstrate an example of task tree retrieval using an example FOON in Figure \ref{fig:tree}. 
An example task tree obtained from this FOON is shown as Figure \ref{fig:stree}.
For emphasis, the graph has been color-coded to distinguish between functional units of two subgraphs; video 1 is denoted by the lime-green object nodes and the red motion nodes,
video 2 is denoted by the dark-green object nodes and the indigo motion nodes, and the overlapped functional units are denoted by the yellow object nodes and orange motion nodes.
The goal is to make cooked ribs (highlighted in purple) using a set of objects observed in the scene (highlighted in blue).
We can start at these blue nodes to arrive to our goal. 
These nodes can be viewed as root nodes in basic tree structures, except that a tree does not typically have multiple roots.
The path we take is entirely dependent on the availability of the objects in the robot's environment.
The ability to merge and combine knowledge into one single network makes our network very powerful and useful. Within a universal FOON lies many possible task trees for different scenarios.
These possibilities can be an entirely novel way of executing a task, as there may be several ways of creating a particular meal. Our task sequences therefore will not necessarily
follow the entire procedure from a single video source. For example, there are many ways to prepare a sauce for meat, and by using the knowledge on how to prepare sauces with a
variety of ingredients we can compensate for the unavailability of certain items needed if we instead followed one recipe.
The novelty not only comes from the possibility of different task sequences, but also in the flexibility in how we prepare the meals.





\subsection{Motion Generating}
The task tree is then used to generate a task sequence that contains a series of motions, objects, and their states, which provides step-by-step commands executable by a robot.
After a functional unit in the task tree is provided and the involved objects are identified in an environment, a new trajectory of the motion needs to be generated using the locations
of those objects as constraints. The new trajectory is generated using motion harmonics, given by:

\begin{equation*} \label{eq-gnew}
y_\text{new}(d, t) =
\begin{cases}
\bar{g}(d, t) + c(d)^{\T}g(d, t) + c_0(d) & d \notin D_q \\
\text{tanh}\left(\bar{g}(d, t) + c(d)^{\T}g(d, t) + c_0(d)\right) & d \in D_q
\end{cases}
\end{equation*}
where weights $c(d)$ and $c_0(d)$ are the variables, and $\T$ denotes transpose. Let $D_e$ denote the set of degrees on which constraints are imposed. Let $e_s(d)$ represent the $C$-th constraint on degree $d\in D_e$, and $t_s$ be the time stamp of $e_s(d)$, $s=1, 2, \dots, S$. The new trajectory tries to resemble the demonstrated data as well as meeting the constraints. For $d\notin D_e$,  $c(d) = \mu(d)$ and $c_0(d) = 0$. For $d\in D_e$, we define the loss function as:
\begin{align*}
L=& \frac{1}{2}\sum_s\left(e_s(d) - \bar{g}(d, t_s) - c_0(d) - c(d)^{\T}g(d, t_s) \right)^2 \nonumber \\
& + \frac{\lambda}{2}\left(c(d) - \mu(d)\right)^{\T}\Sigma(d)^{-1}\left(c(d) - \mu(d)\right),
\end{align*}
where $\lambda$ is a hyper-parameter chosen by the user. The optimal weights that minimize $L$ are obtained by first computing $c(d)$ by solving $Ac(d) = b$, where:
\begin{align*}
A &= \sum_sg'(d, t_s)g'(d, t_s)^{\T} + \lambda\Sigma(d)^{-1} \\
b &= \sum_sf_s'(d)g'(d, t_s) + \lambda\Sigma(d)^{-1}\mu(d)
\end{align*}

\hspace{0.3cm}\mbox{$g'(d, t_s) \overset{\text{def}}{=} g(d, t_s) - \frac{1}{S}\sum_sg(d, t_s)$}, 

\hspace{0.7cm}\mbox{$f'_s(d) \overset{\text{def}}{=} f_s(d) - \frac{1}{S}\sum_sf_s(d)$}, and

\hspace{0.7cm}\mbox{$f_s(d) \overset{\text{def}}{=} e_s(d) - \bar{g}(d, t_s)$}
\newline

\noindent and then use $c(d)$ to compute $c_0(d)$:
\begin{equation*}
c_0(d) = \frac{1}{S}\left(\sum_sf_s(d) - c(d)^{\T}\sum_sg(d, t_s) \right).
\end{equation*}

\begin{figure}
	\includegraphics[width=\linewidth]{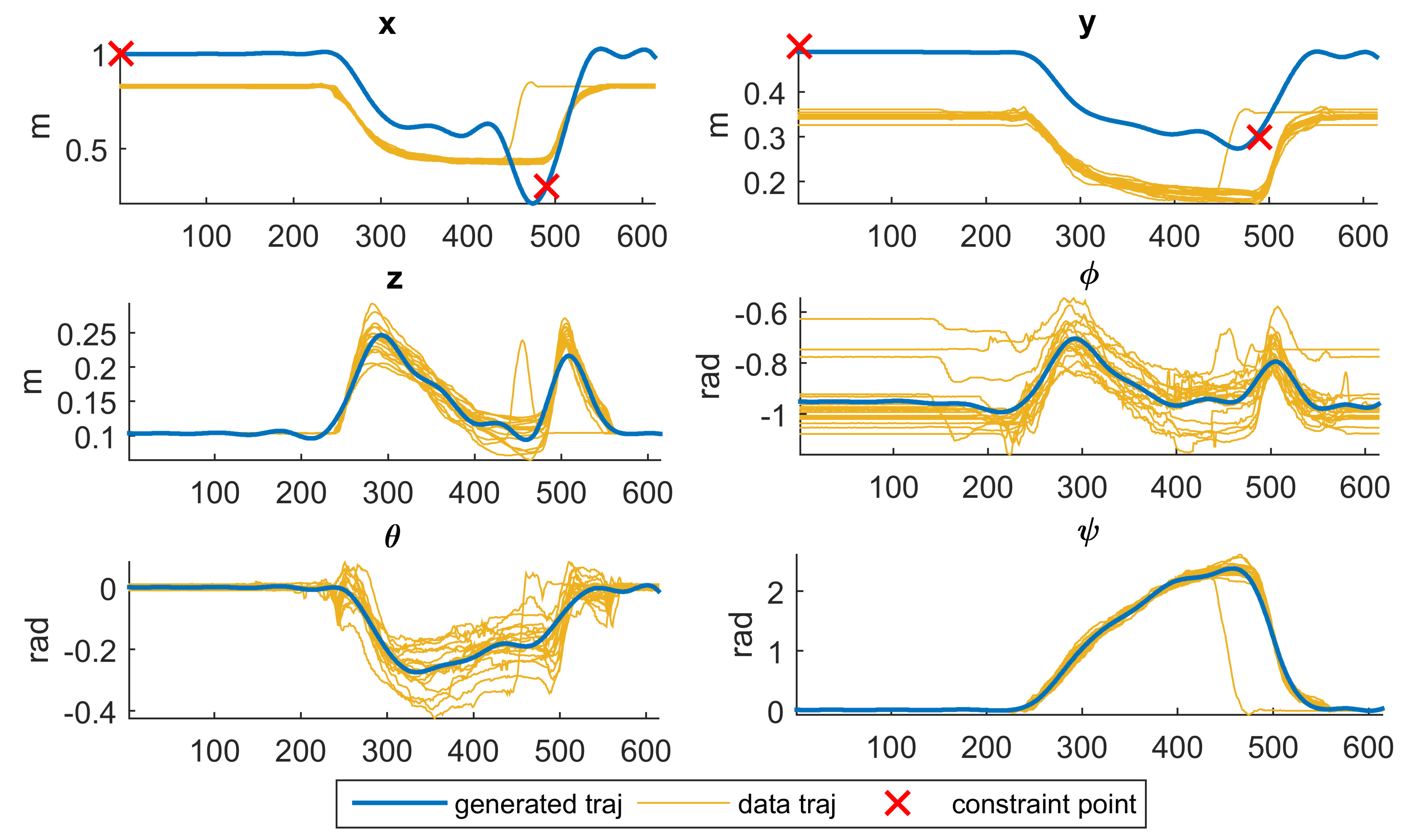}
	\caption{In degree $x$ and $y$, the new trajectory meets the constraints well. Without constraints, the rest degrees of the new trajectory equal the mean of the data. The `data traj' (in dark yellow) have been aligned using DTW.}
	\label{fig-motion}
\end{figure}

Figure \ref{fig-motion} shows one example: a generated pouring motion trajectory in a new environment with a new relative start and target positions between a cup and a teapot. The pouring motion was learned from twenty trials of pouring instances. According to the new relative position of pair, we applied two constraints to $x$ and $y$, which represent the start and target of the new pouring motion: $e_1 = [1, 0.5]^{\T}$, $e_2 = [0.3, 0.3]^{\T}$, $t_1 = 1$, $t_2 = 490$. We set $\lambda=1e-6$. From the results in Figure \ref{fig-motion}, we can see that the generated trajectory resembles the learning and satisfies the new constraints. Currently, our motion generation process does not consider dynamic or kinematic constraints of the actual robot.

We developed a simulated kitchen environment using Unity to demonstrate the results of the task sequences and the motion generation approach. 
We have generated several manipulation simulations and attached two simulation videos with this paper.




\section{Conclusion and Future Work}

In this paper, we present the functional object-oriented network or FOON as a representation of manipulation tasks, which connects interactive objects with their functional motions.
A FOON provides structured knowledge about the relationship between the object states and functional object motions, which is valuable for not only learning manipulation tasks but also for understanding human activities.
We developed an approach to construct functional units using abstracted knowledge of online instructional videos, mainly cooking videos; this information is extracted by human users tasked with annotating these functional units manually. 
The functional units are then connected into subgraphs and then merged into a universal FOON through an automatic process. 
A large universal FOON is constructed from 60 online videos, which has been made available online. It has been analyzed to obtain insights of the structure of the network using centrality measures.
Manipulation knowledge can be retrieved from a FOON, given a manipulation goal using our searching algorithm.
The manipulation knowledge is stored in a task tree sequence with a series of involved objects, manipulation motions, and immediate goals.
These task trees will not necessarily follow the same exact procedure as described by a single recipe or video, making them a flexible and novel way of manipulating objects based on the knowledge acquired from several sources.

The motion nodes in FOON are described as a combination of motion harmonics. 
With the parameterized representation, a new motion of a learned type can be generated to accommodate new constraints in different environments and motion contexts. Based on the obtained task tree, a sequence of manipulation motions is generated properly to perform the desired task. 
In addition to the examples illustrated in the paper, generated manipulation motion sequences are demonstrated in a simulated kitchen environment and two demo videos are attached with this paper, and additional demo videos are available at \cite{foonet}.



In the future, we plan to perform more network analyses, such as computing connection strength and efficiency, on our universal FOON to better understand the dynamics of the network.
Additionally, we hope to integrate probabilities into our representation using the findings we obtained from the analyses we have done as well.
We are also exploring means of using object similarity to obtain more practical solutions through inference.
For instance, even though we have not seen how certain objects are used or manipulated, we can instead use the knowledge we know and apply it to an unknown problem.
Missing information such as quantities will also be considered.
We are also seeking methods to potentially solve the problem of automatically generating a FOON from instructional videos. 


\section*{Acknowledgement}
This material is based upon work supported by the National Science Foundation under Grant No. 1421418.


\bibliographystyle{unsrt}
\bibliography{ref}

\end{document}